%% file: root.tex
\documentclass[letterpaper, 10 pt, conference]{ieeeconf}  

\IEEEoverridecommandlockouts                              

\input{include/binbin.tex}

\usepackage{graphics} 
\usepackage{epsfig} 
\usepackage{mathptmx} 
\usepackage{times} 
\usepackage{amsmath} 
\usepackage{amssymb}  

\input{include/notation-math-defs.tex}
\usepackage{multirow}
\usepackage{hyperref}

\newcommand{\paper}{paper}

\usepackage[capitalise,noabbrev]{cleveref}
\title{\LARGE \bf
	Learning to Complete Object Shapes for Object-level Mapping in Dynamic Scenes
}

\author{Binbin Xu$^{1}$, Andrew J. Davison$^{1}$, Stefan Leutenegger$^{1,2}$
\thanks{1 The authors are with Department of Computing, Imperial College London, United Kingdom.
 	{\tt\small \{b.xu17, a.davison, s.leutenegger\}@imperial.ac.uk}
 }
\thanks{2 The author is also with the Smart Robotics Lab, Technical University of Munich, Germany 
{\tt\small \{stefan.leutenegger\}@tum.de}
}
\thanks{The supplementary video can be watched on: https://youtu.be/mH22H7jp1D8}
}

\begin{document}

\maketitle
\thispagestyle{empty}
\pagestyle{empty}

\begin{abstract}
\input{include/abstract.tex}
\end{abstract}

\input{include/main.tex}

\bibliographystyle{IEEEtran}
\bibliography{robotvision}

\end{document}


\maketitle
\thispagestyle{empty}
\pagestyle{empty}


The learnable network parameters in this work include three parts, canonical correspondence network $F_n$, shape prior network $F_0$, and shape posterior network $F_1$.

\begin{figure}[t]
	\centering
	\includegraphics[width=\linewidth]{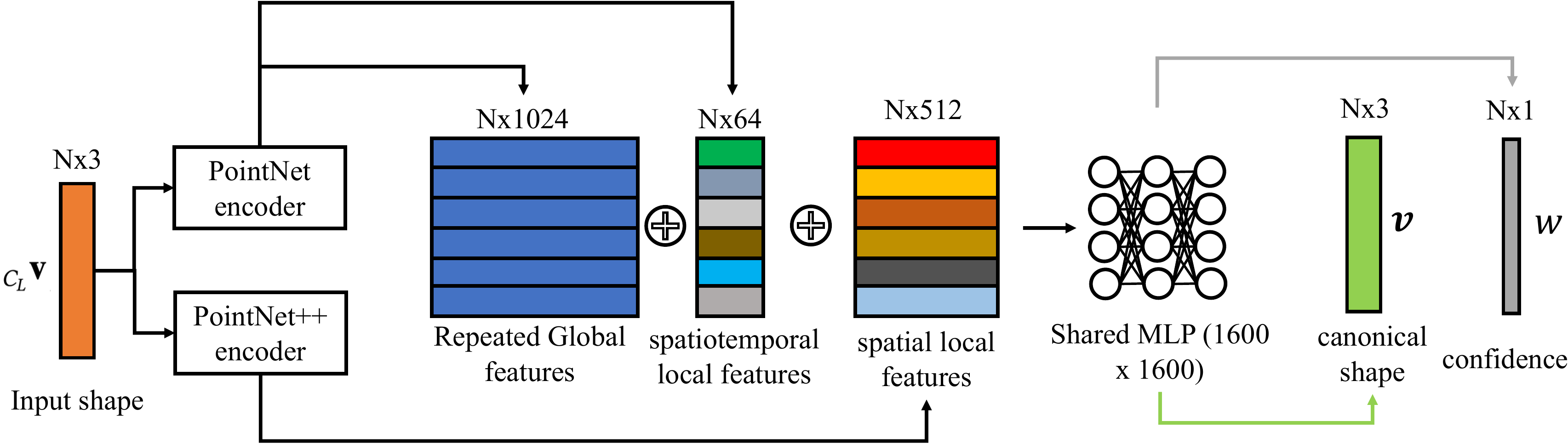}%
	\caption{The architecture of our canonical correspondence network. The architecture is modified from \cite{Rempe:etal:NIPS2020}. It extracts global features and spatiotemporal local features from the PointNet encoder~\cite{Qi:etal:CVPR2017} and spatial local features from the PointNet++ encoder~\cite{Qi:etal:NIPS2017}. These features are concatenated and passed to an MLP to regress the canonical shape correspondence and the associated confidence.}
	\label{fig:ch5_nocs_network}

\end{figure}

\cref{fig:ch5_nocs_network} shows the architecture of our canonical correspondence network $F_n$. It takes the partial pointcloud $\generalThree{C_L}{v}{}$ from the depth measurements as input and predicts its correspondence $\mbf v$ in canonical space and the associated confidence $w$.
We train the canonical correspondence network using partial pointclouds generated from the synthetic shapenet dataset~\cite{Shapenet:ARXIV2015}. During the training, we augment the input pointcloud with random object poses and solve the 7DoF object poses using \cref{eq: init loss}. To help network prediction robust to outliers, we also add random depth outliers in the pointcloud generation to learn the correspondence confidence $w$ in a self-supervised way. The solved pose is compared to the augmented ground-truth pose and the whole network is trained end-to-end since the estimation is differentiable. 

\begin{equation}
	\label{eq: init loss}
	\argmin_{s_{O_n}, \T{C_L}{O_n}} \sum_{\pixel{L} \in M_n} w \left( \mbf v - \frac{1}{s_{O_n}}\T{C_L}{O_n}^{-1} \generalThree{C_L}{v}{} (\pixel{L}) \right).
\end{equation}

\begin{figure}[t]
	\centering
	\includegraphics[width=0.8\linewidth]{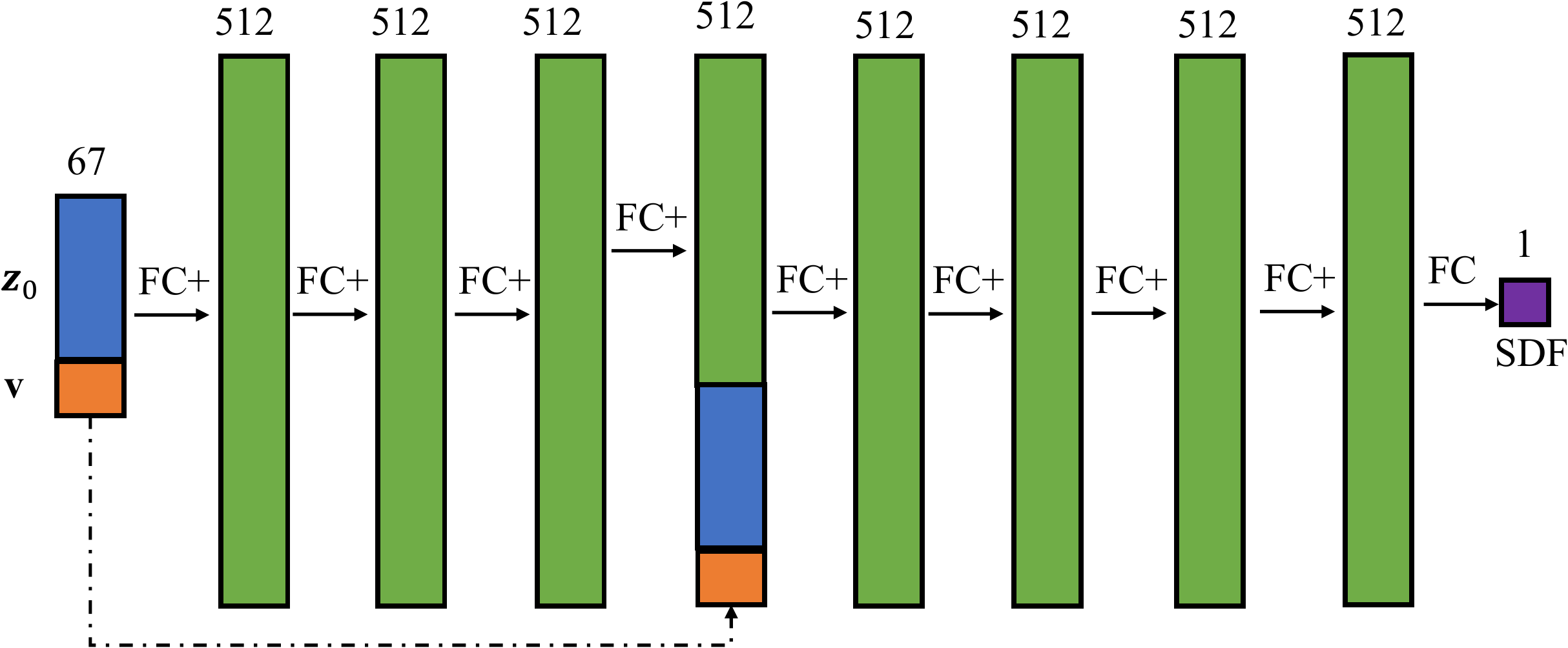}%
	\caption{The architecture of the shape prior network~\cite{Park:etal:CVPR2019}. The input vector is fed through a decoder, which contains eight fully-connected (FC) layers with one skip connection. FC+ denotes a FC with a following softplus activation and the last FC layer output a single SDF value.}
	\label{fig:ch5_deepsdf}
\end{figure}

\begin{figure}[t]
	\centering
	\includegraphics[width=\linewidth]{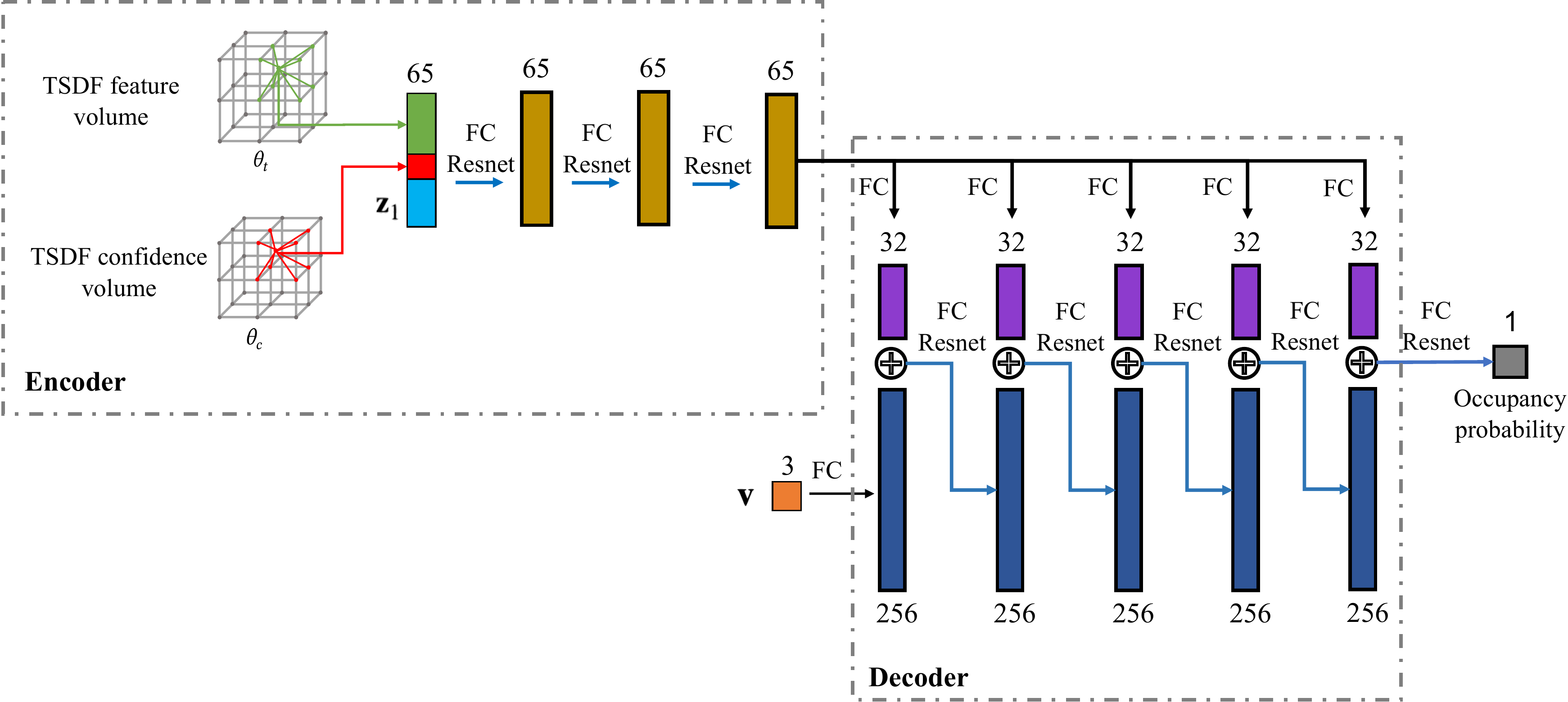}%
	\caption{The architecture of our shape completion network, modified from CONet~\cite{Peng:etal:ECCV2020}. The encoder extracts the TSDF feature vector $\theta_t[\mbf v] \in \mathbb{R}^{32} $ and the TSDF confidence vector $\theta_c[\mbf v] \in \mathbb{R}^{1}$ from TSDF feature volume and TSDF confidence volume, respectively, and concatenates them with a latent code $z_1$ as an input to the network. It goes through 3 fully-connected (FC) ResNet-blocks to extract local latent features, which are then fed into an occupancy decoder~\cite{Mescheder:etal:CVPR2019} to predict occupancy probabilities on the position vector $\mbf v$. }
	\label{fig:ch5_shape_completion_net}
\end{figure}

We use the pre-trained off-shelf network weights from the category-level shape prior network DeepSDF \cite{Park:etal:CVPR2019} for $F_0$, which was also trained in the synthetic shapenet dataset~\cite{Shapenet:ARXIV2015}. Its architecture is visualized in \cref{fig:ch5_deepsdf}.

\begin{figure}[t]
	\centering
	\includegraphics[width=\linewidth]{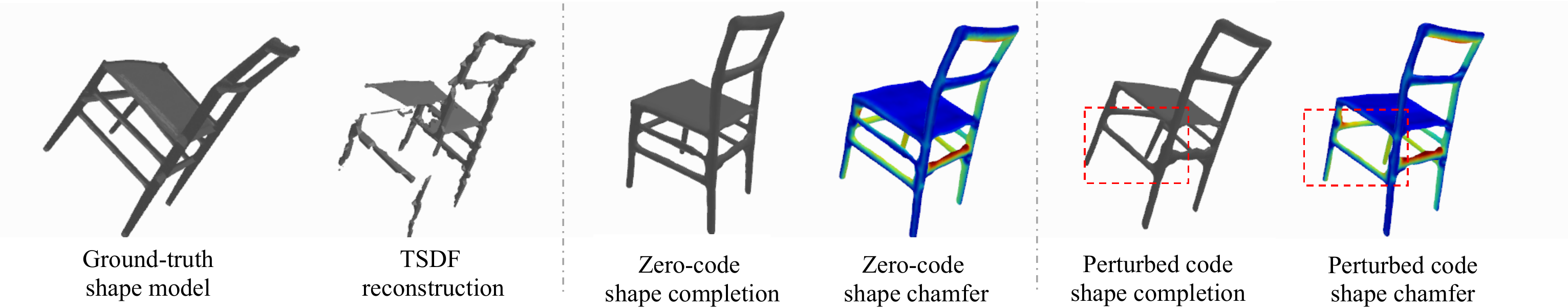}%
	\caption{Editing the conditioned latent code can change the geometry of the unobserved part in the object model.}
	\label{fig:opt_latent}
\end{figure}

\cref{fig:ch5_shape_completion_net} shows the architecture of our posterior shape completion network $F_1$. It takes the input of a TSDF feature volume and a TSDF confidence volume, which are extracted separately from a partial TSDF volume and its weight volume. The input of TSDF confidence volume is designed to balance the observed depth measurement and shape prior information. The unseen part has TSDF weight of zero value and biases towards prior shape and will gradually switch to 3D reconstruction when more depth information is fused into the corresponding TSDF voxel. We additionally concatenate the inputs with a latent code $\mbf z_1 \in \mathbb{R}^{32}$ so that the hidden space can be optimised to generate novel shapes, as shown in \cref{fig:opt_latent}. The shape completion network $F_1$ can predict a complete object geometry represented in a continuous occupancy function by inferring an occupancy value $o$ on any given 3D position $\mbf v \in \mathbb R^3$ in canonical space.

Since the partial observation in reality mostly happens due to self-occlusions and sometimes also due to occlusions from other objects, we rendered depth maps using objects in the shapenet dataset~\cite{Shapenet:ARXIV2015}. To train the posterior shape completion network, we rendered depth maps for each object in the shapenet dataset~\cite{Shapenet:ARXIV2015} to simulate partial depth observations.
We use the occupancy loss defined in \cref{eq: occupied} to encourage the completed shape to be similar to the ground-truth one. 
\begin{equation}
	\label{eq: occupied}
	E_\mathrm{occ} = -\sum_{{\mbf v}} [o_{\mbf v} \log(o_{\mbf v}^*) + (1-o_{\mbf v}) \log(1-o_{\mbf v}^*)].
\end{equation}
Similar to the training in DeepSDF~\cite{Park:etal:CVPR2019}, different object shapes belonging to the same category have different latent codes, but share the same decoder network weight. We make different partial observations of the same object shape share the same latent code.

\bibliographystyle{IEEEtran}
\bibliography{robotvision}

%% file: include/binbin.tex
\usepackage{graphicx}
\usepackage{subfloat}
\usepackage[caption=false]{subfig}
\usepackage{tikz}

\usepackage{tabularx}
\usepackage{multirow}
\usepackage{booktabs}	
 
\usepackage{color}

\usepackage{amssymb,amsmath}
\usepackage{amsfonts}

%% file: include/notation-math-defs.tex
\usepackage{xifthen}
\usepackage{xparse}
\usepackage{subdepth}
\usepackage{leftidx}
\usepackage{bm}
\usepackage{amssymb,amsmath}
\usepackage{amsfonts}

\newcommand{\bbm}{\begin{bmatrix}}
	\newcommand{\ebm}{\end{bmatrix}}

\DeclareMathAlphabet{\mbf}{OT1}{ptm}{b}{n}
\newcommand{\mbs}[1]{{\bm{#1}}}

\newcommand{\mbsbar}[1]{{\overline{\boldsymbol{#1}}}}
\newcommand{\mbshat}[1]{{\hat{\boldsymbol{#1}}}}
\newcommand{\mbstilde}[1]{{\tilde{\boldsymbol{#1}}}}

\newcommand{\mbsdot}[1]{{\dot {\boldsymbol{#1}}}}

\newcommand{\mbfbar}[1]{{\overline{\mbf{#1}}}}
\newcommand{\mbfhat}[1]{{\hat{\mbf{#1}}}}
\newcommand{\mbftilde}[1]{{\tilde{\mbf{#1}}}}

\newcommand{\mbfdot}[1]{{\dot{\mbf{#1}}}}

\newcommand{\cframe}[1]{{\smash{\protect\underrightarrow{\mathcal{F}}_{#1}}}}

\DeclareMathAlphabet{\mathbfit}{OML}{cmm}{b}{it}
\newcommand{\homo}[1]{{\mathbfit{#1}}}

\newcommand{\mbfh}[1]{{\homo{#1}}}

\newcommand{\argmin}{\operatornamewithlimits{argmin}}

\newcommand{\trans}[3]{\leftidx{_{#1}}{\mbf r}{\IfValueTF{#2}{_{#2#3\hspace{2pt}}}{}}} 

\newcommand{\vel}[3]{\leftidx{_{#1}}{\mbf v}{\IfValueTF{#2}{_{#2#3\hspace{2pt}}}{}}} 
\newcommand{\veltilde}[3]{\leftidx{_{#1}}{\mbftilde v}{\IfValueTF{#2}{_{#2#3\hspace{2pt}}}{}}} 
\newcommand{\velbar}[3]{\leftidx{_{#1}}{\mbfbar v}{\IfValueTF{#2}{_{#2#3\hspace{2pt}}}{}}} 
\newcommand{\velhat}[3]{\leftidx{_{#1}}{\mbfhat v}{\IfValueTF{#2}{_{#2#3\hspace{2pt}}}{}}} 
\newcommand{\veldot}[3]{\leftidx{_{#1}}{\mbfdot v}{\IfValueTF{#2}{_{#2#3\hspace{2pt}}}{}}} 

\newcommand{\acc}[3]{\leftidx{_{#1}}{\mbf a}{\IfValueTF{#2}{_{#2#3\hspace{2pt}}}{}}} 
\newcommand{\acctilde}[3]{\leftidx{_{#1}}{\mbftilde a}{\IfValueTF{#2}{_{#2#3\hspace{2pt}}}{}}} 
\newcommand{\accbar}[3]{\leftidx{_{#1}}{\mbfbar a}{\IfValueTF{#2}{_{#2#3\hspace{2pt}}}{}}} 

\newcommand{\rotvel}[3]{\leftidx{_{#1}}{\mbs \omega}{\IfValueTF{#2}{_{#2#3\hspace{2pt}}}{}}} 
\newcommand{\rotveltilde}[3]{\leftidx{_{#1}}{\mbstilde \omega}{\IfValueTF{#2}{_{#2#3\hspace{2pt}}}{}}} 
\newcommand{\rotvelbar}[3]{\leftidx{_{#1}}{\mbsbar \omega}{\IfValueTF{#2}{_{#2#3\hspace{2pt}}}{}}} 
\newcommand{\rotvelhat}[3]{\leftidx{_{#1}}{\mbshat \omega}{\IfValueTF{#2}{_{#2#3\hspace{2pt}}}{}}} 
\newcommand{\rotveldot}[3]{\leftidx{_{#1}}{\mbsdot \omega}{\IfValueTF{#2}{_{#2#3\hspace{2pt}}}{}}} 

\newcommand{\Crot}[2]{\leftidx{}{\mbf C}{_{#1#2\hspace{2pt}}}} 
\newcommand{\T}[2]{\leftidx{}{\mbfh T}{_{#1#2\hspace{2pt}}}} 

\newcommand{\generalThree}[3]{\leftidx{_{#1}}{\mbf #2}{_{#3}}} 
\newcommand{\pixel}[1]{{\mbfh u}{_{#1}}} 

%% file: include/abstract.tex
In this paper, we propose a novel object-level mapping system that can simultaneously segment, track, and reconstruct objects in dynamic scenes. It can further predict and complete their full geometries by conditioning on reconstructions from depth inputs and a category-level shape prior with the aim that completed object geometry leads to better object reconstruction and tracking accuracy. For each incoming RGB-D frame, we perform instance segmentation to detect objects and build data associations between the detection and the existing object maps. A new object map will be created for each unmatched detection. For each matched object, we jointly optimise its pose and latent geometry representations using geometric residual and differential rendering residual towards its shape prior and completed geometry. 
Our approach shows better tracking and reconstruction performance compared to methods using traditional volumetric mapping or learned shape prior approaches. We evaluate its effectiveness by quantitatively and qualitatively testing it in both synthetic and real-world sequences.

%% file: include/main.tex
\section{Introduction}
Simultaneous Localisation and Mapping (SLAM) research aims to concurrently estimate both the scene geometry of the unknown environment as well as the robot pose within it from the data of its on-board sensors only. It has rapidly progressed from sparse SLAM~\cite{Davison:etal:PAMI2007, Klein:Murray:ISMAR2007} into dense SLAM~\cite{Newcombe:etal:ISMAR2011, Vespa:etal:RAL2018}, and recently into semantic object-level SLAM~\cite{McCormac:etal:ICRA2017, McCormac:etal:3DV2018}. This fast-evolving research has enabled many robotic applications. Despite this, most SLAM research still assumes a static scene, where points in the 3D world maintain constant spatial positions in a global coordinate. Any information violating this assumption, such as moving objects in the environment, would be treated as outliers and are intentionally ignored in tracking and mapping steps.

This setup, however, can only handle a small amount of dynamic elements, excluding itself from many real-world applications as environments, particularly where humans are involved, are continually changing. A robust SLAM system capable of handling highly dynamic environments, therefore, is desirable. Most current dynamic SLAM research can be categorised into three main directions. One maps the whole changing world in a non-rigid deformable representation to deal with the changing topology of deformable/moving objects~\cite{Newcombe:etal:CVPR2015}. The second aims at improving the robustness and accuracy of camera tracking by ignoring all possibly moving objects and building a single static background model~\cite{Jaimez:etal:ICRA2017, Scona:etal:ICRA2018, Bescos:etal:RAL2018}. The third models dynamic environments by creating object-centric maps for each possibly moving rigid object in the scene while fusing corresponding information into these object-level maps~\cite{Runz:Agapito:ISMAR2018, Xu:etal:ICRA2019}. Object-level tracking and mapping can be conducted for each object and camera motion against the static part of the map. This paper aligns with the last direction as we believe that, similar to human perception, an instance-awareness of the surrounding environment can help intelligent robots perceive the scene changes and enables meaningful interactions with the surrounding environment.

\begin{figure}[t]
	\centering
	\includegraphics[width=\linewidth]
	{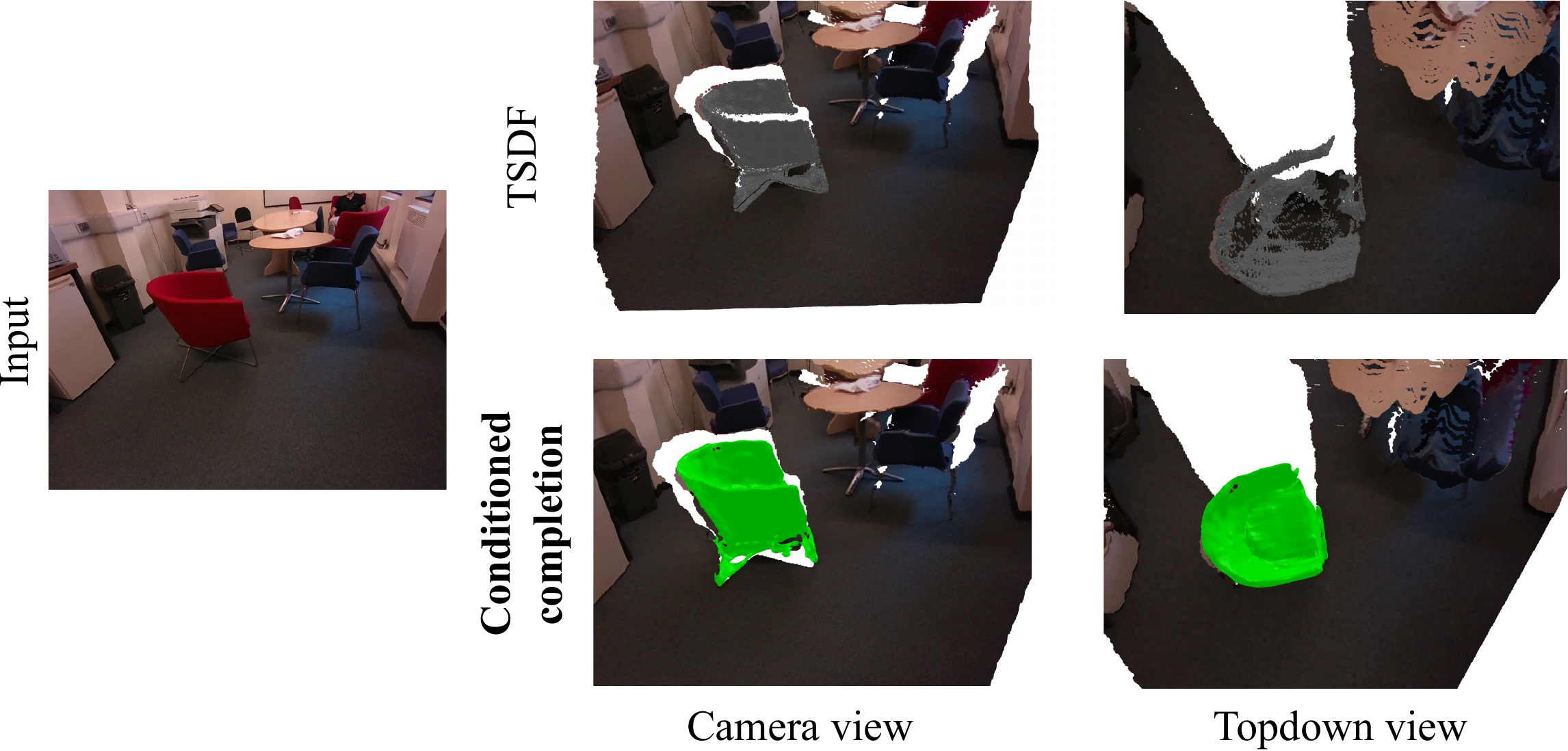}
	\caption{Given RGB-D images, our system builds object-level dense dynamic maps that can robustly track camera pose and object poses while completing the missing sensor information using object priors. Compared to the classic TSDF maps, our object maps fill in unobserved parts and their latent codes can be optimised jointly with object poses. Interfered regions by humans can be detected and intentionally removed in the system. The background pointclouds are projected for pure visualisation purpose.}
	\label{fig:teaser_completion}
\end{figure}

By far most existing object-level dynamic SLAM systems mentioned above adopt the classic map representation that have been exploited in the static SLAM systems, such as pointclouds~\cite{Bescos:etal:RAL2021}, surfels~\cite{Runz:Agapito:ISMAR2018} or volumetric maps~\cite{Xu:etal:ICRA2019}. This leads to partial or incomplete object maps as only the observable information can be fused into the object models. Information in unseen parts can not be filled unless an object or the sensor is moved actively. Contrary to reconstructing objects from scratch, some works recently explored learning-based category-level object shape priors and build object-level maps based on learned shape priors~\cite{Sucar:etal:3DV2020, Wang:etal:3DV2021}. The object geometry and pose are usually optimised via differentiable rendering. However, most of these systems are only applicable in static scenes. Besides, despite being able to generate complete object geometry, object shape priors cannot capture complex geometry details as the bottleneck of its latent representation can only interpolate shapes inside the training datasets~\cite{Park:etal:CVPR2019}. When combined with dense image alignment, such as photometric or ICP residuals, this inconsistency between the measurement and the object prior model inevitably leads to inaccurate object motion trajectory estimates.

This work stands in the middle between reconstructing object geometry from scratch and mapping using object shape priors. We reconstruct the observable part by continuously fusing depth measurements into a volumetric canonical space and predict the complete geometry by conditioning it on the fused volume. The resulting object geometry can preserve the details that have been observed in the past and simultaneously complete the missing geometry information. We also verified that this completed object geometry can further improve the accuracy of object tracking. The main contributions in this \paper~can be summarised as follows:
\begin{enumerate}
\item we present, to the best of our knowledge, the first RGB-D object-level dynamic mapping system that can complete unseen parts of objects using a shape prior encoded in neural networks while still reconstructing observed parts accurately;
\item a joint optimisation of object pose and shape geometry based on geometric residuals and differentiable rendering;
\item extensive experiments of object tracking and reconstruction components on synthetic and real-world data to evaluate the benefits of object geometry completion for object-level SLAM.
\end{enumerate}


\section{Related Works}
\paragraph{Object-level dynamic SLAM}
Although object-level dynamic SLAM research can be dated back to \cite{Wang:etal:ICRA2003}, visual dense object-level dynamic SLAM has only been explored recently. From RGB-D sensor inputs, Co-Fusion~\cite{Runz::Agapito::ICRA2017} segments objects by either ICP motion segmentation or semantic segmentation and then tracks objects separately based on ElasticFusion~\cite{Whelan:etal:IJRR2016}. MaskFusion~\cite{Runz:Agapito:ISMAR2018} segments objects using a combination of instance segmentation from Mask-RCNN and geometric edges, and tracks objects using the same approach as Co-Fusion. Both Co-Fusion and MaskFusion use surfels to represent map models, which is memory efficient but cannot directly provide free space information in the map, and neither surface connectivity. DynSLAM-II~\cite{Bescos:etal:RAL2021} extends from ORB-SLAM II~\cite{Mur-Artal:etal:TRO2017} and formulates object tracking using sparse feature descriptor matching. Object maps are represented using clusters of pointclouds, which can bring object poses and geometries into the pose graph optimisation but also lack space connection awareness. 
Instead, MID-Fusion~\cite{Xu:etal:ICRA2019} uses memory-efficient octree-based volumetric signed distance field (SDF) representation for objects and re-parametrises tracking residuals in object coordinates and weights. EM-Fusion~\cite{Strecke:Stuckler:CVPR2019} similarly uses volumetric SDF object maps but formulates object tracking as direct alignment of input frames with the SDF representations. Their following work \cite{Strecke:Stuckler:CVPR2020} infers the missing object geometry by penalising the hull and intersection constraints. However, it did not explore shape prior information and requires heavy computation to optimise SDF field explicitly. Instead, we fuse the depth measurement into object-level SDF maps and predict completed object geometries by incorporating a shape prior in continuous occupancy fields.

\paragraph{SLAM with shape prior maps}
Instead of estimating both object geometry and poses from scratch, some approaches use a shape prior to represent objects. Since the coordinates of object shape priors and the run-time measurement are not necessarily aligned, a relative rigid transformation needs to be estimated. This is analogous to the localisation-only problem in SLAM. SLAM++~\cite{Salas-Moreno:etal:CVPR2013} is one of the pioneering object-level RGB-D SLAM systems. It scanned objects in advance and then maps the detected instances at run-time by jointly optimising a pose graph of camera and object poses. Relying on pre-scanned objects, however, limits its ability to scale to unknown object models. Rather than using instance-level shape priors, several following works exploited category-level shape priors as there is limited variance in certain object categories. The category-level shape prior can be learnt in various representations, such as PCA models as in DirectShape~\cite{Wang:etal:ICRA2020}, occupancy grids as in Deep-SLAM++ \cite{Hu:etal:ARXIV2019}, variational autoencoders as in NodeSLAM~\cite{Sucar:etal:3DV2020}, or autodecoders as in DSP-SLAM~\cite{Wang:etal:3DV2021}.  However, most of these works only target static environments, as multi-view consistency of static world points is required to localise the shape prior models. \cite{Li:etal:RAL2021} relax this restriction using a Bayesian filter to associate object detections on different frames and fuse the prior model by simply averaging the latent codes from each frame. However, object shape deviations cannot always be captured by the shape prior interpolation. The object tracking accuracy would be affected by the discrepancy between the prior shape model and the online measurement. 
We address this challenge by conditioning the completion network on the integrated 3D reconstruction model.

\paragraph{Object-level tracking}
To track moving objects in RGB-D sequences, several pioneering object-level works adopt the frame-to-model tracking methods from RGB-D SLAM systems \cite{Runz::Agapito::ICRA2017} and parametrise them for object tracking~\cite{Xu:etal:ICRA2019, Bescos:etal:RAL2021}. The classic photometric residual, however, is difficult to deal with as object lighting changes; several approaches explore using learning-based robust features to formulate object tracking in direct \cite{Xu:etal:RAL2021} and in-direct ways \cite{Wen:Bekris:IROS2021}. Parallel to learning category-level shape priors, Wang et al.~\cite{Wang:etal:CVPR2019N} proposed to learn category-level pixel-wise correspondence from RGB-D images to the canonical space. The shape is implicitly defined from this correspondence and the frame-to-canonical transformation can be estimated from this noisy correspondence. Rempe et al.~\cite{Rempe:etal:NIPS2020} further proposed to generate more stable correspondences by accumulating temporal information from RGB-D sequences. Recently, Muller et al.~\cite{Muller:etal:CVPR2021} proposed to track moving objects and predict their complete geometry using such canonical correspondence representation. In this work, the object pose is initially predicted using canonical correspondence regression. However, we found it does not necessarily yield alignment to the canonical space and we further optimise the pose tracking using geometric residual and differential rendering.

\begin{figure*}[tbp]
	\centering
	\includegraphics[width=0.7\linewidth]{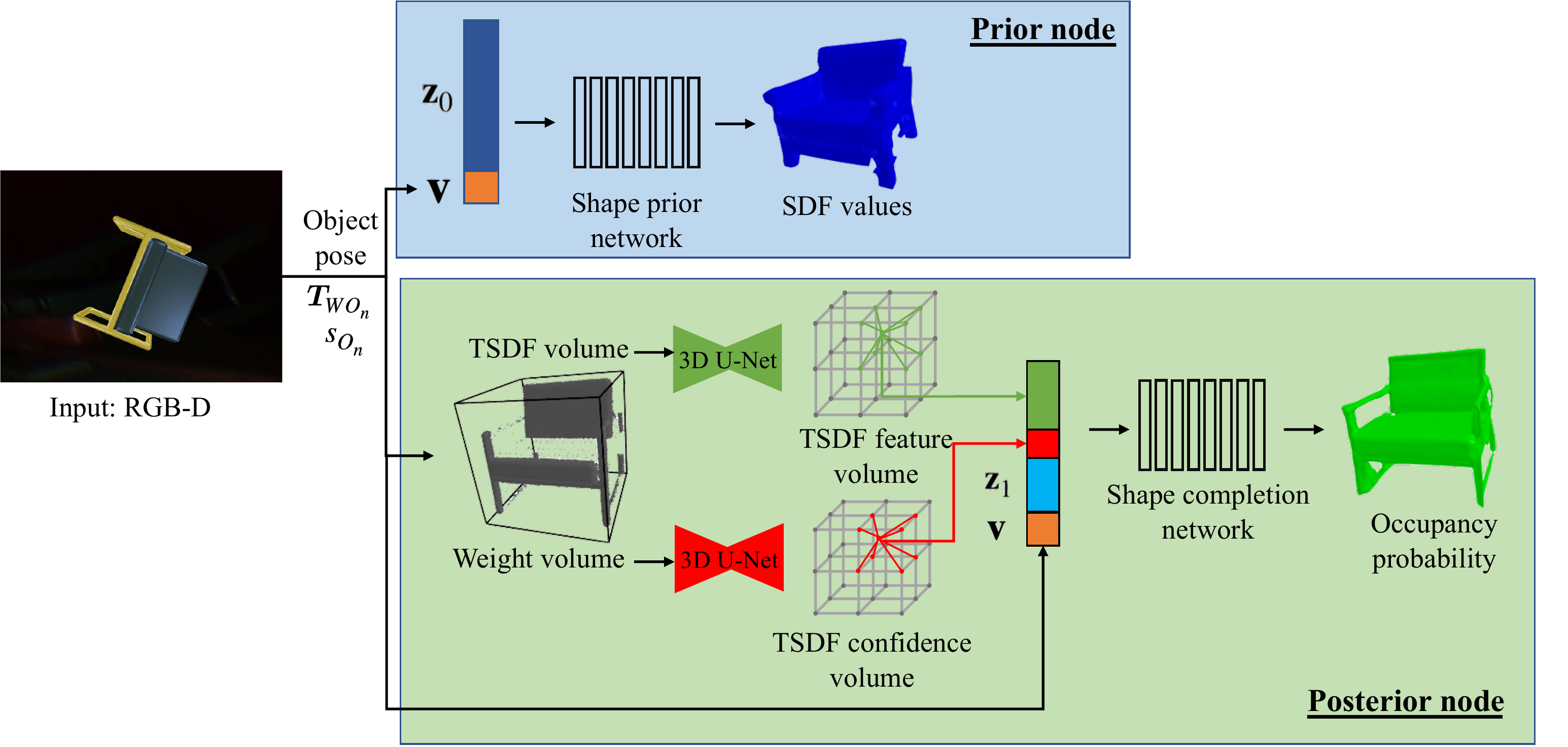}
	\caption{The overview of our object geometry representation.}
	\label{fig:mapping_overview}
\end{figure*} 

\begin{figure}[tbp]
	\centering
	\includegraphics[width=\linewidth]{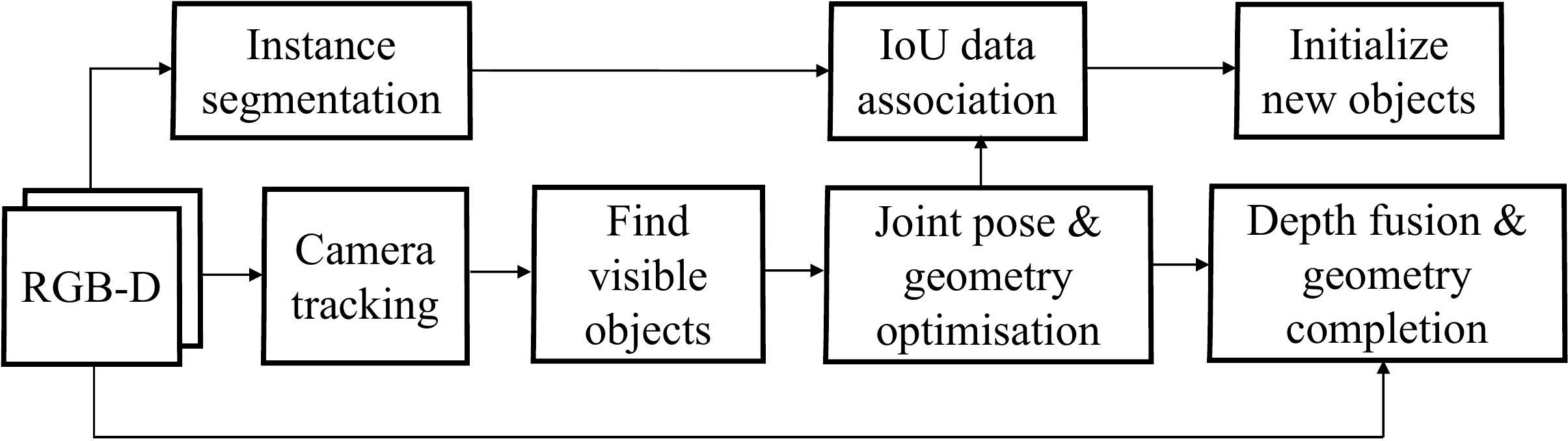}
	\caption{The pipeline of the proposed method}
	\label{fig:pipeline}
\end{figure}

\section{Method}
\subsection{Notations and Preliminaries}
In this work, we will use the following notation: a coordinate frame is denoted as $\cframe{A}$.  The homogeneous transformation from $\cframe{B}$ to $\cframe{A}$ is denoted as $\T{A}{B}$, which is composed of a rotation matrix $\Crot{A}{B}$ and a translation vector $\trans{A}{A}{B}$. 

Every detected object is represented in its individual object coordinate frame $\cframe{O_n}$, with $n \in \{ 0\ldots, N\}$, where $N$ is the total number of objects (excluding background) and 0 denotes background. We assume a canonical static volumetric model is stored in each object coordinate frame, forming the basis of our multi-instance SLAM system. To leverage the category-level shape prior, we need to align the canonical space with the one defined in training, otherwise the completion performance will be deteriorated since it cannot fully take advantage of the shape variances of the objects in the same category. The relative transformation between the current world coordinate and the corresponding object canonical space is defined as a joint state composed of a rigid transformation $\T{W}{O_n}$ and the object scale $s_{O_n}$. We define the object pose using this joint state. $\T{W}{O_n}$ needs to be continuously updated for a moving object but the object scale should be consistent across different frames. 

\subsection{System Overview}
\cref{fig:pipeline} shows the pipeline of our proposed system. 
Each input RGB-D image is processed by Mask R-CNN to perform instance segmentation.
The camera pose is tracked against background regions, excluding the human mask area and moving objects, similar to what has been proposed in MID-Fusion~\cite{Xu:etal:ICRA2019}.

The object geometry is composed of two nodes with a shared object pose: prior node and posterior node, as shown in \cref{fig:mapping_overview}. The prior node represents its category-level shape prior using a latent code $\mbf z_0 \in \mathbb{R}^{64}$. It can be used to express the continuous SDF field $s$ on any query 3D location in object canonical coordinate $\mbf v \in \mathbb R^3$  using a DeepSDF shape prior network $F_0$~\cite{Park:etal:CVPR2019} as 
\begin{equation}
\label{eq:deepsdf}
s = F_0(\mbf v, \mbf z_0).
\end{equation}
The prior node is used to initialise the object pose and re-localise the object model when object tracking is lost as its shape is not affected by measurements. The posterior node encodes a fused partial TSDF volume and its associated TSDF weight volume into TSDF feature volume $\theta_t$ and TSDF confidence volume $\theta_c$ separately using 3D-UNet~\cite{Cciccek:etal:MICCAI2016}. Then a complete occupancy field $o$ can be predicted on any given 3D position $\mbf v \in \mathbb R^3$ using a shape completion network $F_1$:
\begin{equation}
\label{eq:conv-onet}
o = F_1(\mbf v, \theta_t[\mbf v], \theta_c[\mbf v], \mbf z_1).
\end{equation}
where $\theta_t[\mbf v] \in \mathbb{R}^{32} $ and $\theta_c[\mbf v] \in \mathbb{R}^{1}$ denote the feature vectors tri-linearly interpolated at the point $\mbf v$ inside the volumes $\theta_t$ and $\theta_c$, respectively.
We additionally condition it on a latent code $\mbf z_1 \in \mathbb{R}^{32}$ so that the hidden space can be optimised to generate novel shapes. 
The shape completion network $F_1$ shares a similar architecture to the CONet~\cite{Peng:etal:ECCV2020}, but also takes extra inputs of TSDF confidence weight and an instance-level latent code.
 

We perform an efficient Axis Aligned Bounding Box (AABB) ray intersection test~\cite{Majercik:etal:JCGT2018} to find all the visible existing object models in the current viewpoint and render object masks for each visible models. An Intersection of Union (IoU) between the detections on the current frame and the rendered model masks is computed to build data associations between the current frame detections and existing object models. Then we track each object model and complete its geometry by performing a joint optimisation of pose and geometry (\cref{subsec: joint_opt}).  Using estimated poses of the camera and objects, new depth measurements are fused into an object model and a complete shape geometry can be predicted by conditioning on the fused model. 
New objects are created for unmatched detections by initialising their initial object pose using object prior models.

\subsection{Joint Optimisation of Object Pose and Geometry}
\label{subsec: joint_opt}
Instead of defining an arbitrary canonical space for object coordinates~\cite{Xu:etal:ICRA2019}, we need to estimate the 7DoF relative transformation, which is composed of $\T{W}{O_n}$ and $s_{O_n}$, to align the initialised object coordinate to the training canonical space for each object detection in order to take advantage of the learned prior information.

\paragraph{Initialisation}
\label{method:init}
Given an RGB-D frame $(I_L, D_L)$ and detected object mask $M_n$, we predict their positions in the canonical space $\mbf v$ and associated confidences $w$ from back-projected pointcloud using a modified normalised object correspondence network $F_n$ from  \cite{Rempe:etal:NIPS2020}:
\begin{equation}
	\generalThree{C_L}{v}{}(\pixel{L}) =  \pi^{-1}(\pixel{L}, D_L[\pixel{L}]), \forall \pixel{L} \in M_n,
\end{equation}
\begin{equation}
	\label{eq: nocs_regression}
	F_n\left( \generalThree{C_L}{v}{} \right) \rightarrow \mbf v, w
\end{equation}

Then we solve the 7-DoF relative transformation, scale $s_{O_n}$, rotation $\Crot{C_L}{O_n}$, and translation $\trans{C_L}{C_L}{O_n}$ from the regressed correspondences using the Umeyama algorithm~\cite{Ummenhofer:PAMI1991} with SVD decomposition: 
\begin{equation}
	\label{eq: init loss}
	\argmin_{s_{O_n}, \T{C_L}{O_n}} \sum_{\pixel{L} \in M_n} w \left( \mbf v - \frac{1}{s_{O_n}}\T{C_L}{O_n}^{-1} \generalThree{C_L}{v}{} (\pixel{L}) \right).
\end{equation}

For the latent codes $\mbf z_0$ and $\mbf z_1$, we use both zero code initialisations.
We only run this pose initialisation step for new unmatched object detections. The initial pose solved from SVD decomposition, however, is necessary to be close to the ground-truth canonical pose, due to the unseen shapes or viewpoints, affecting the performance of shape completion.

\paragraph{Coarse Estimation}
\label{method:coarse}
To refine object canonical poses from the initial pose prediction, we jointly optimise it with the prior latent code $\mbf z_0$ to minimise the 3D SDF loss $E_\mathrm{SDF}$ and 2D rendering loss $E_\mathrm{render}$:
\begin{equation}
	\label{eq: coarse_loss}
	E_\mathrm{coarse} = \lambda_\mathrm{s} E_\mathrm{SDF}+ \lambda_\mathrm{r0} E_\mathrm{render} + \lambda_\mathrm{z0} ||\mbf z_0||.
\end{equation}
The 3D SDF loss is defined to encourage the back-project depth points to align with the object surface, where the zero SDF value is defined
\begin{equation}
	\label{eq: sdf_loss}
	E_\mathrm{SDF} = \sum_{\pixel{L} \in M_n} F_0 \left( \frac{1}{s_{O_n}} \T{C_L}{O_n}^{-1} \generalThree{C_L}{v}{} (\pixel{L}), \mbf z_0 \right).
\end{equation}

We cannot compute SDF residuals for empty space since ground-truth SDF values are not available at test time. Instead, for the non-surface areas, we use differentiable rendering to encourage the rendered depth $D_L$ to be close to the measured depth $\tilde D_L$. We compute the rendering loss for the visible 3D space inside the object 3D bounding box: 
\begin{equation}
	\label{eq: render_loss}
	E_\mathrm{render} = \sum_{\pixel{L} \in B_n} D_L[\pixel{L}] - \tilde D_L[\pixel{L}],
\end{equation}
where
\begin{equation}
D_L[\pixel{L}] = \sum_{i=1}^N w_i d_i,
\end{equation}
and $w_i$ is the ray-termination probability \cite{Sucar:etal:3DV2020} of sample $i$ at depth $d_i$ along the ray from the pixel $\pixel{L}$,
\begin{equation}
	w_i = o_i \prod_{j=1}^{i-1} (1-o_j),
\end{equation}
and $B_n$ is the 2D bounding box rendering from the estimated object 3D bounding box on this frame.
A continuous occupancy field can be extracted from the continuous prior SDF field as proposed in \cite{Wang:etal:3DV2021}:
\begin{equation}
	\label{eq: sdf_to_occ}
	o_i = -\frac{1}{2\sigma} F_0 \left( \frac{1}{s_{O_n}} \T{C_L}{O_n}^{-1} \left( \pi^{-1}(\pixel{L}, d_i) \right), \mbf z_0 \right),
\end{equation}
where $\sigma$ is the truncation distance to control the transition.

By freezing the network weight in $F_n$, the cost function in \cref{eq: coarse_loss} can be iteratively solved using Gauss-Newton optimisation with analytical Jacobians. Since the prior shape is not necessarily aligned with the actual observation, it is unnecessary to sample every pixel ray. Instead, we run this optimisation on sparse ray samples, which can speed up the optimisation without losing much accuracy. 

\paragraph{Dense Refinement}
\label{method:dense}
After the coarse estimation, we have a rough alignment of the object model with the canonical space. However, the optimised object prior $\mbf z_0$ cannot necessarily capture the details of depth measurements. To further align the object model and to complete the hidden part, we jointly optimise the posterior latent code $\mbf z_1$ with the object pose by minimizing a 3D occupancy loss $E_\mathrm{occ}$ on both occupied and empty space (excluding the unknown 3D space) and a similar 2D rendering loss $E_\mathrm{render}$:
\begin{equation}
	\label{eq: fine loss}
	E_\mathrm{refine} = \lambda_\mathrm{o} E_\mathrm{occ} + \lambda_\mathrm{r1} E_\mathrm{render} + \lambda_\mathrm{z1} ||\mbf z_1||.
\end{equation}

The occupied space is defined on the back-projected points and the empty space is uniformly sampled in the background space as well as the foreground space before the depth measurement. The occupancy loss is defined using the binary cross-entropy loss between the predicted occupancy value $o_{\mbf v}$ from the shape completion network using \cref{eq:conv-onet} and the ground-truth occupancy values $o_{\mbf v}^*$ (0.5 for the occupied space and 0 for the empty space) for sampled points $\mbf v$ inside the occupied and empty space:
\begin{equation}
	\label{eq: occupied}
	E_\mathrm{occ} = -\sum_{{\mbf v}} [o_{\mbf v} \log(o_{\mbf v}^*) + (1-o_{\mbf v}) \log(1-o_{\mbf v}^*)].
\end{equation}

Similar to the coarse estimation, we can also evaluate the 2D rendering loss using \cref{eq: render_loss}. The difference is here we use the decoded continuous occupancy value for the sampled $d_i$ using \cref{eq:conv-onet}, instead of converting it from the SDF field in \cref{eq: sdf_to_occ}. The refined object pose $\T{C_L}{O_n}$ is used to integrate the current depth frame into the corresponding TSDF volume and weight volume \cite{Newcombe:etal:ISMAR2011}.

\subsection{Training Setup}
The learnable network parameters in this work includes three part, canonical correspondence network $F_n$, shape prior network $F_0$, shape posterior network $F_1$.

We train the canonical correspondence network using the partial pointcloud generated from the synthetic shapenet dataset~\cite{Shapenet:ARXIV2015}. During training, we augmented the input pointcloud with random object pose and solve the 7DoF object poses using \cref{eq: init loss}. To help network prediction robust to outliers, we also added random depth outliers in the pointcloud generation to learn the correspondence confidence $w$ in a self-supervised way. The solved pose is compared to the augmented ground-truth pose and the whole network is trained end-to-end since the estimation is differentiable. 

We use the pre-trained off-shelf network weights for the category-level shape prior network $F_0$ \cite{Park:etal:CVPR2019}, which was also trained in the shapenet dataset~\cite{Shapenet:ARXIV2015}. To train the posterior shape completion network, we rendered depth maps for each object in the shapenet dataset~\cite{Shapenet:ARXIV2015} and trained the shape completion network using the partial depth observation. We use the occupancy loss defined in \cref{eq: occupied} to encourage the completed shape to be similar to the ground-truth one. Similar to the latent code training in DeepSDF~\cite{Park:etal:CVPR2019}, different object shapes have their own latent codes, which are trained together with the network. We make different partial observations of the same object shape share the same latent code.

\section{Experiments}
\subsection{Quantitative Reconstruction Evaluations}
\subsubsection{Experimental Setup}
We validate the reconstruction quality of our method on object-level surface reconstruction tasks. We conduct a comparison on the chair category of the ShapeNet~\cite{Shapenet:ARXIV2015} dataset. The split of train/val/test sets follows the same setting in \cite{Peng:etal:ECCV2020}. We randomly select 50 models from the test set to conduct quantitative evaluations. We generate input depth images by rendering images using uniformly sampled virtual camera viewpoints surrounding the CAD model. The hyperparameters used in inference optimization are chosen as $\sigma = 0.05$, $\lambda_\mathrm{s} = 100$, $\lambda_\mathrm{r0} = 2.5$, $\lambda_\mathrm{z0} = 5$, $\lambda_\mathrm{o} = 100$, $\lambda_\mathrm{r1} = 1$, and $\lambda_\mathrm{z1} = 1$.

\subsubsection{Baseline Methods}
To evaluate the object mapping, we compare with the following baseline methods:
\begin{itemize}
	\item TSDF-fusion: We fuse the depth measurements into a TSDF volume grid as in \cite{Newcombe:etal:ISMAR2011}.
	\item DeepSDF mapping: We use the pre-trained decoder weight in \cite{Park:etal:CVPR2019}. As the shape completion code is not provided, we optimise the SDF loss on the input pointcloud as well as the empty space constraint proposed in IGR~\cite{Gropp:etal:ICML2020}.
	\item CONet: We use the weights trained from partial pointcloud input in \cite{Peng:etal:ECCV2020} and pass the accumulated pointcloud in the canonical space to generate the continuous occupancy field where the meshes are extracted.
\end{itemize}

\subsubsection{Metrics}
To quantitatively evaluate the quality and completeness of the shape reconstruction, we use the following metrics:
\begin{itemize}
	\item IoU: 
	We sample 100k points uniformly in the bounding box and evaluate on both the reconstructed and the ground-truth meshes whether each point is inside or outside. The final value is the fraction of intersection over union. Higher is better.
	\item Chamfer Distance: 
	we sample 100k points on the surface of both the ground-truth and the reconstructed mesh. We compute the closest points from the reconstructed to the ground-truth mesh using kD-tree and vice-versa. We then compute the average of the L1 distances to the closest points in each direction. Lower is better.
	\item (In-)completeness: As the completeness of the object map is essential in this work, we also report completeness, which is the one-way chamfer distance from the ground-truth meshes to the reconstructed ones. This is to measure the closest distance from each ground-truth mesh points to the reconstruction. Lower is better.
\end{itemize}

\subsubsection{Results and Discussions}
\begin{figure}[htb]
	\centering
	\subfloat[][IoU] 
	{\includegraphics[width=0.48\linewidth]{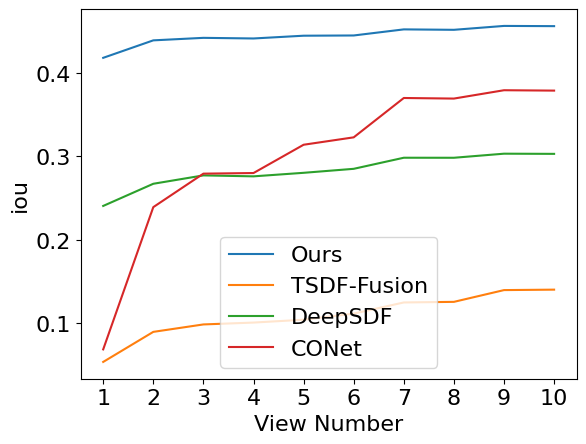}
	}
	\hfill
	\subfloat[][Chamfer distance (L1)] 
	{\includegraphics[width=0.48\linewidth]{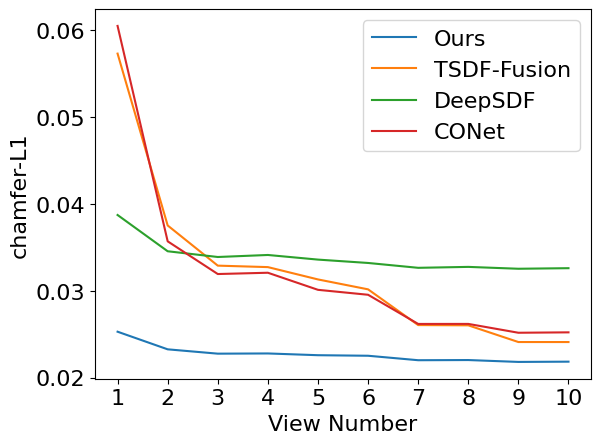}
	} \\
	\subfloat[][(In-)Completeness (lower is better)] 
	{\includegraphics[width=0.48\linewidth]{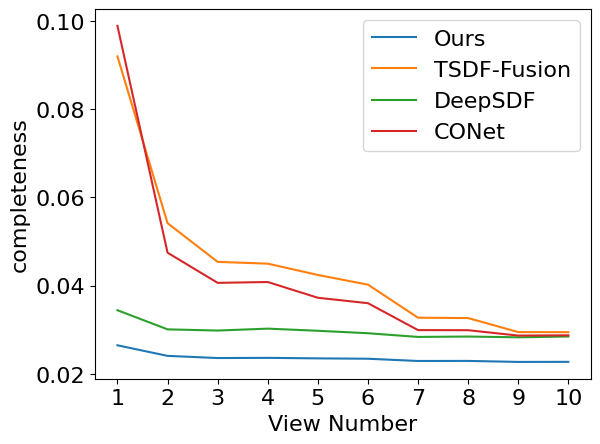}
	} \\
	\caption{Quantitative comparison of reconstruction quality and completion of our proposed methods v.s.\ classic TSDF-Fusion, learning-based DeepSDF and CONet. Our proposed method consistently show better reconstruction results from single view depth completion to multiple views.}
	\label{fig:eval_reconstruction}
	
\end{figure}

\begin{figure}[htb]
	\centering
	\includegraphics[width=0.9\linewidth]{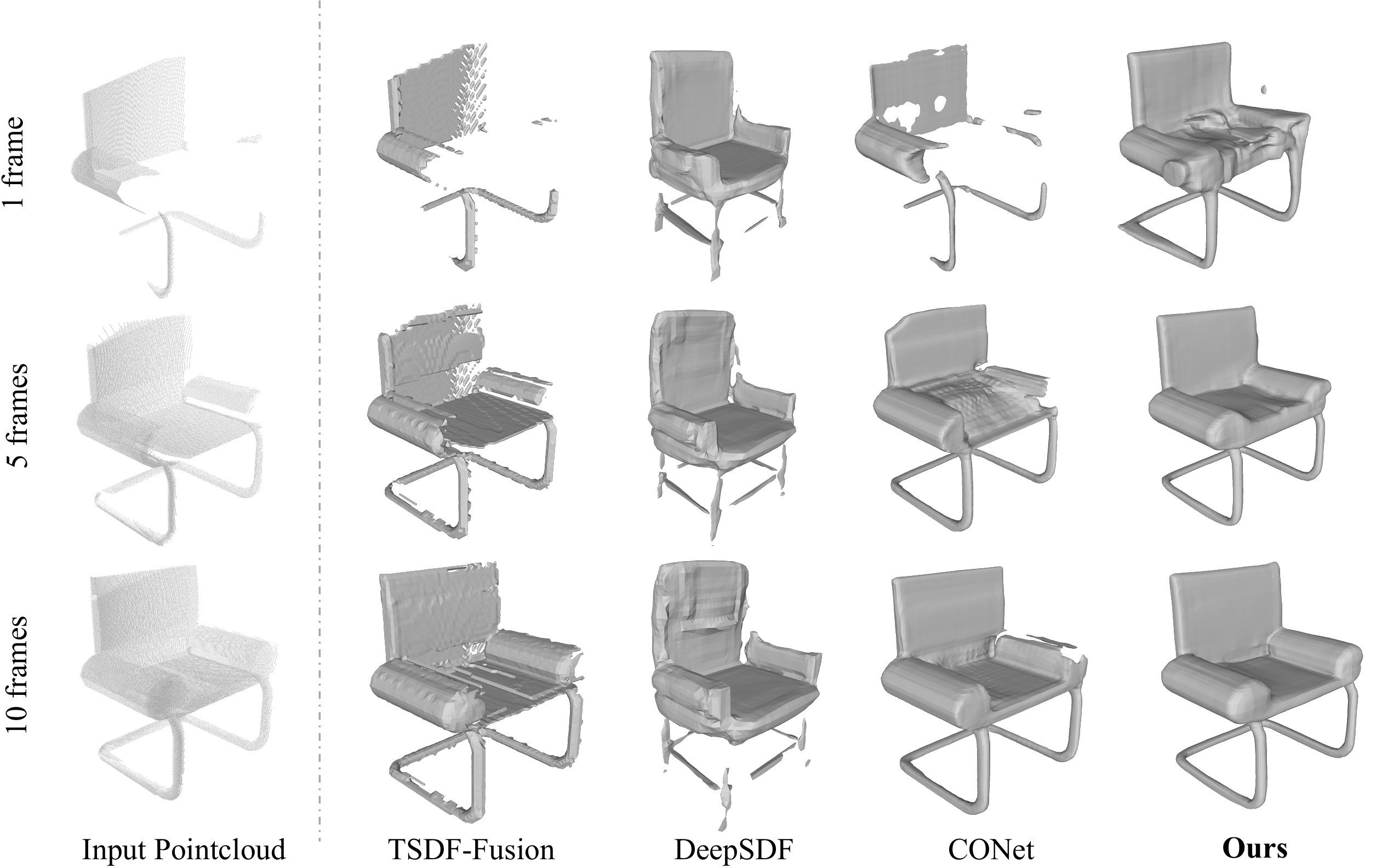}%
	\caption{Qualitative Results on reconstructions. Our method is superior to all other methods in completing missing information and reconstructing fine details.}
	\label{fig:eval_map_vis}
\end{figure}

We quantitatively evaluate how the view number of depth measurement would impact the reconstruction results of different methods. \cref{fig:eval_reconstruction} reports the result. It can be seen that our proposed method consistently show better reconstruction results from single view depth completion to multiple views. When the view number is limited, classic TSDF-Fusion shows worse result as it can only reconstruct the visible parts. CONet completes some missing information, but still struggles as it heavily depends on the input pointcloud. DeepSDF does not condition on the input and the latent code optimisation can fit the few depth measurement and shows better completion and reconstruction results. Our proposed method uses both the input information and shape prior information, yielding best performance. When more depth measurements are received, TSDF-Fusion and CONet start to fill in the missing information while DeepSDF struggles to leverage more measurement information. Our result also improves since we can also take advantage of the measurement information. \cref{fig:eval_map_vis} shows an examples of reconstruction results by each method in the view number case of 1, 5, and 10.

%

\subsection{Quantitative Tracking Evaluations}
\subsubsection{Experimental Setup}
To quantitatively evaluate the object-level tracking and mapping performance, we randomly select 10 object models from the test split of the chair category in the ShapeNet~\cite{Shapenet:ARXIV2015} and render 200 frames using Blender. To ensure diversity of object motion, texture, and illuminations, we randomise four point light sources, camera viewpoint, and object trajectories. We then subsample the sequences using sampling intervals {1, 2, 4} in order to create small, medium and large motion subsets.

\subsubsection{Baseline Methods}
To evaluate the object tracking, we compare with the following baseline methods:
\begin{itemize}
	\item RGB-D VO: a non-learning-based visual odometry method proposed in \cite{Steinbrucker:etal:ICCVW2011}, which minimises the photometric loss between two frames. We re-parametrised it for object tracking.
	\item Point-to-Plane ICP: a non-learning geometric registration method \cite{Rusinkiewicz:Levoy:3DMIN2001}
	\item Color ICP: a non-learning registration method using both color and geometric information \cite{Park:etal:ICCV2017}
	\item Prior: a state-of-the-art object pose estimation using DeepSDF shape prior model. It is originally proposed in \cite{Wang:etal:3DV2021} for static object pose estimation and we re-parametrised it for estimating moving objects. This is equivalent to an ablation study of using the prior model only without conditioned completion refinement from Section III-C c).
	\item NOCS: a state-of-the-art learning-based canonical correspondence regression method. We adopted the network architecture proposed in \cite{Rempe:etal:NIPS2020}.  This is equivalent to an ablation study of using only the initial prediction from Section III-C a).
\end{itemize}

\subsubsection{Metrics}
To quantitatively evaluate the accuracy of the object tracking, we use the following metrics:
\begin{itemize}
	\item ATE: Absolute. Trajectory Error defined in \cite{Sturm:etal:IROS2012}, in the unit of of meters
	\item RPE\_t: relative pose error (RPE) metrics in translation defined in \cite{Sturm:etal:IROS2012}, in the unit of of metres
	\item RPE\_R: relative pose error (RPE) metrics in rotation defined in \cite{Sturm:etal:IROS2012}, in the unit of of degrees
	\item R\_err: mean orientation error on each frame individually, in the unit of degrees
	\item t\_err: mean translation error on each frame individually, in the unit of metres
\end{itemize}
The above metrics all indicate better tracking performance when the values are lower. To analyse the trajectory, we align the first frame of the estimated object pose to the ground-truth canonical space.

\subsubsection{Results and Discussions}
\cref{table: quan-track} reports the experimental results. It shows that our approach consistently outperforms both the learning-based and non-learning-based methods in small and large motion situations. For non-learning approaches, RGB-D VO~\cite{Steinbrucker:etal:ICCVW2011}, Point-to-Plane ICP~\cite{Rusinkiewicz:Levoy:3DMIN2001}, and Color ICP~\cite{Park:etal:ICCV2017} only leverage the depth and intensity information from two-view measurements, without taking into account any object shape prior information. The single view canonical correspondence prediction from NOCS~\cite{Rempe:etal:NIPS2020} only considers shape prior information and does not take advantage of the multiview constraint. Our proposed method instead combines both multi-view constraint and shape prior information into object pose estimation for better tracking accuracy. Similar to ours, the shape prior method~\cite{Wang:etal:3DV2021} adopts category-level shape prior from DeepSDF~\cite{Park:etal:CVPR2019} and uses differential rendering to estimate object poses. However, latent code optimisation cannot necessarily capture the geometry deviation between training space and test shapes and thus affects the accuracy of pose estimation. 

\begin{table}[tb]
\centering
\setlength{\tabcolsep}{2.5pt}
\subfloat[][Keyframe gap-1]{
\label{table: track_kf1}
\begin{tabular}{lcccccc}
\toprule
method [unit] &   ATE [m] &  RPE\_t [m] &  RPE\_R [$^{\circ}$] &  R\_err [$^{\circ}$] &  t\_err [m]\\
\midrule
\textbf{Ours} & \textbf{0.030} &  \textbf{0.027} &  \textbf{3.845} &  \textbf{3.931} & \textbf{0.040}  \\
Prior & 0.044 &  0.047 &  8.200 &        6.269 &      0.068  \\
RGBD & 0.254 &  0.106 &  18.47 &       32.25 &      0.314  \\
Point2Plane & 0.047 &  0.035 &  4.672 &        5.970 &      0.064  \\
Color ICP & 0.254 &  0.114 &  29.12 &       56.18 &      0.320  \\
NOCS &    0.074 &      0.059 &        23.46 &       37.87 &      0.085 \\
\bottomrule
\end{tabular}
}
\hfill
\subfloat[][Keyframe gap-2]{
\label{table: track_kf2}
\begin{tabular}{lcccccc}
	\toprule
method [unit] &   ATE [m] &  RPE\_t [m] &  RPE\_R [$^{\circ}$] &  R\_err [$^{\circ}$] &  t\_err [m]\\
\midrule
	\textbf{Ours} &    \textbf{0.033}& \textbf{0.032} & \textbf{5.243} &  \textbf{4.224} & \textbf{0.043}  \\
	Prior &    0.046 &      0.052 &        11.91 &        8.063 &      0.063  \\
	RGBD &    1.068 &      0.403 &        30.89 &       50.95 &      1.309  \\
	Point2Plane &    0.070 &      0.056 &        8.570 &        9.384 &      0.087  \\
	Color ICP &    0.536 &      0.351 &        36.69 &       60.56 &      0.568  \\
	NOCS &    0.074 &      0.074 &       21.65 &       37.78 &      0.084  \\
	\bottomrule
\end{tabular}
}
\hfill
\subfloat[][Keyframe gap-4]{
\label{table: track_kf4}
\begin{tabular}{lcccccc}
\toprule
method [unit] &   ATE [m] &  RPE\_t [m] &  RPE\_R [$^{\circ}$] &  R\_err [$^{\circ}$] &  t\_err [m]\\
\midrule
\textbf{Ours} & \textbf{0.034} &  \textbf{0.038} &  \textbf{6.767} &  \textbf{4.834} & \textbf{0.044}  \\
Prior & 0.043 &  0.050 &  17.20 &        9.885 &      0.061  \\
RGBD & 1.942 &  0.866 &  43.34 &       68.86 &      2.177  \\
Point2Plane & 0.807 &  0.442 &  18.22 &       20.16 &      0.892  \\
Color ICP & 2.786 &  1.885 &  42.73 &       77.89 &      2.802  \\
NOCS & 0.073 &  0.085 &  26.95 &   35.41  &   0.083 \\
\bottomrule
\end{tabular}
}
\caption{Quantitative evaluation of object tracking method on the synthetic moving objects dataset.}
\label{table: quan-track}
\end{table}

\subsection{Timing analysis}
We implemented our system in PyTorch. The average inference time for a pair of RGB-D image in the resolution of 320 $\times$ 240 is 1.337s on a RTX 3090 platform. A more-detailed breakdown of computation time for each component is shown in \cref{table: time-system-component}. A further breakdown of computation time on tracking components is shown in \cref{table: time-tracking-component}.

We would like to highlight that our current implementation is prototyped in Python. We believe a real-time system can be achieved by exploiting C++ and further GPU parallelisation. 


\begin{table}[tb]
\centering
\subfloat[][System components]{
\label{table: time-system-component} 
\begin{tabular}{c|c|c|c}
	\hline
	Components & Tracking &  Integration & Completion (visualization)  \\ \hline
	Time (s)  &  1.284     &   0.003  &  0.474             \\ \hline
\end{tabular}
}
\hfill
\subfloat[][Object tracking components]{
	\label{table: time-tracking-component} 
	\begin{tabular}{c|c|c|c}
		\hline
		Components & Initialization & Coarse est. & Dense refinement  \\ \hline
		Time (ms)  &  0.643        &   0.150          &   1.129         \\ \hline
	\end{tabular}
}
\caption{Run-time analysis (s)}
\vspace{-2em}
\end{table}

\subsection{Qualitative Evaluations}
\begin{figure}[tb]
	\centering
	\subfloat[][Completion of a red chair] 
	{\includegraphics[width=0.9\linewidth]
		{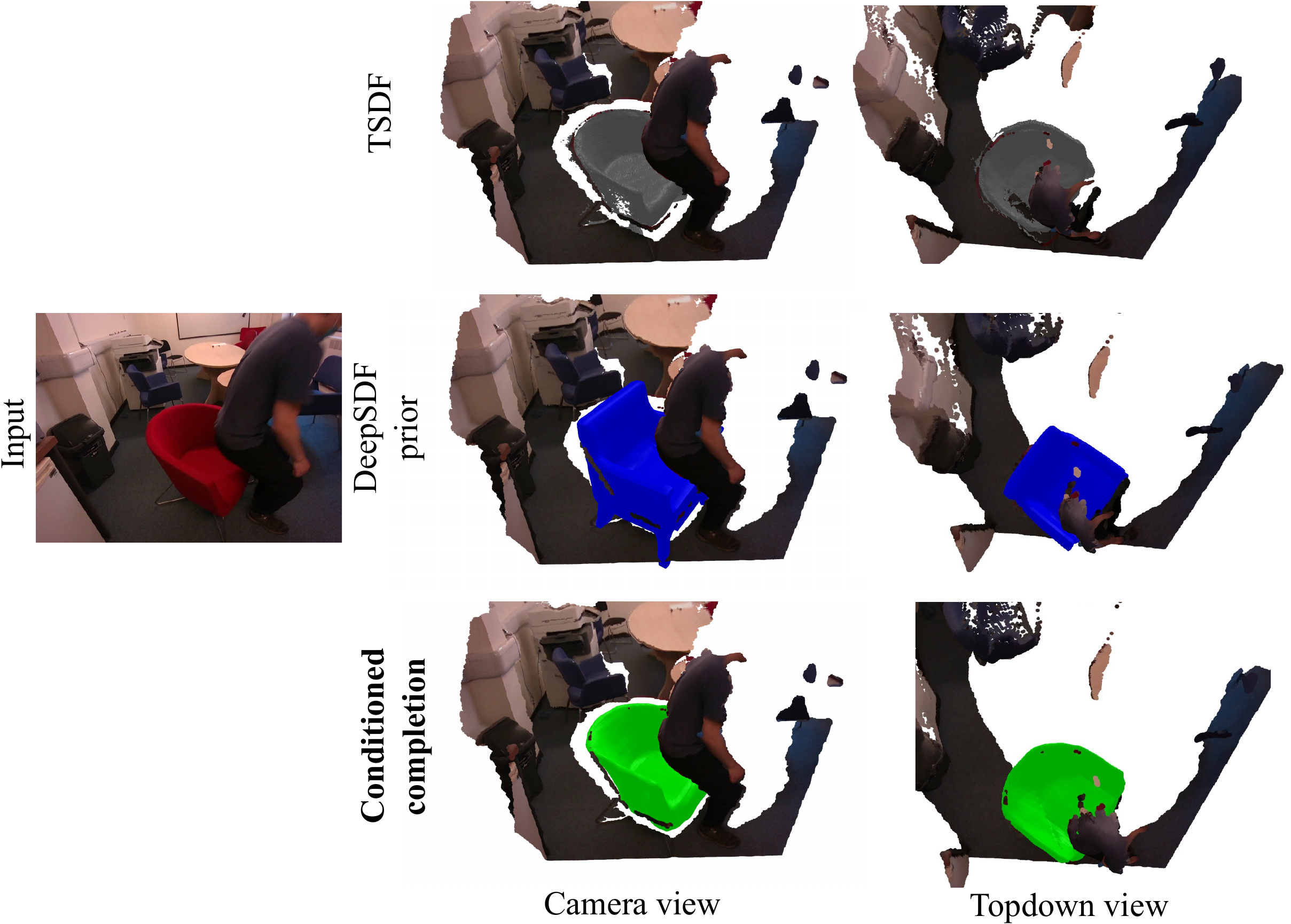}
	}\\
	\subfloat[][Completion of a blue chair] 
	{\includegraphics[width=0.9\linewidth]{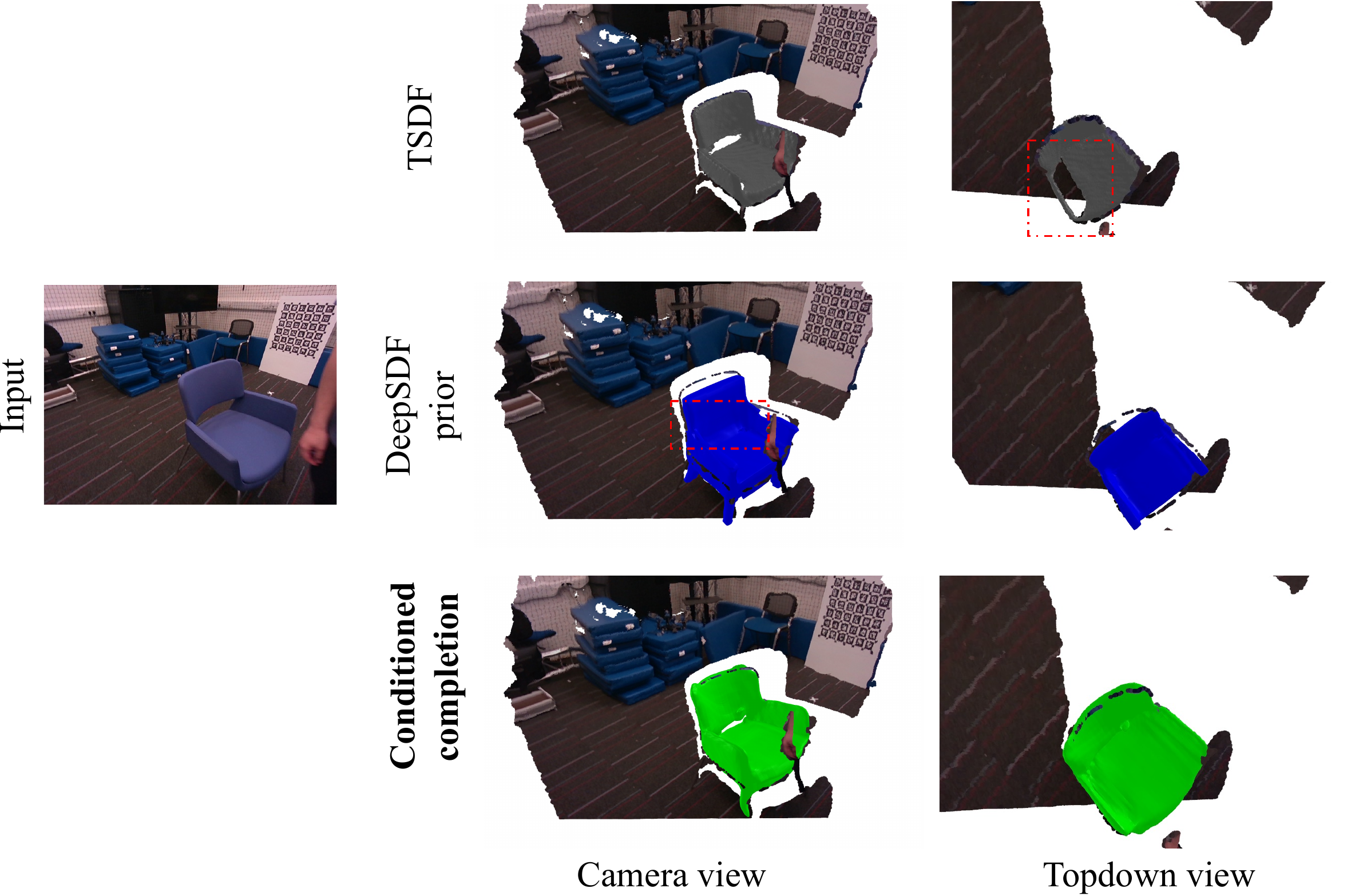}
	} \\
	
	\caption{Qualitative comparison of classic TSDF volume representation (gray), DeepSDF shape prior representation (blue), and our conditioned completion representation (green): our representation can reconstruct the observed part more correctly than a shape prior and completes the unseen part where TSDF fusion fails.}
	\label{fig:shape_completion}
	\vspace{-1em}

\end{figure}

\begin{figure}[tb]
	\centering
	\includegraphics[width=0.9\linewidth]{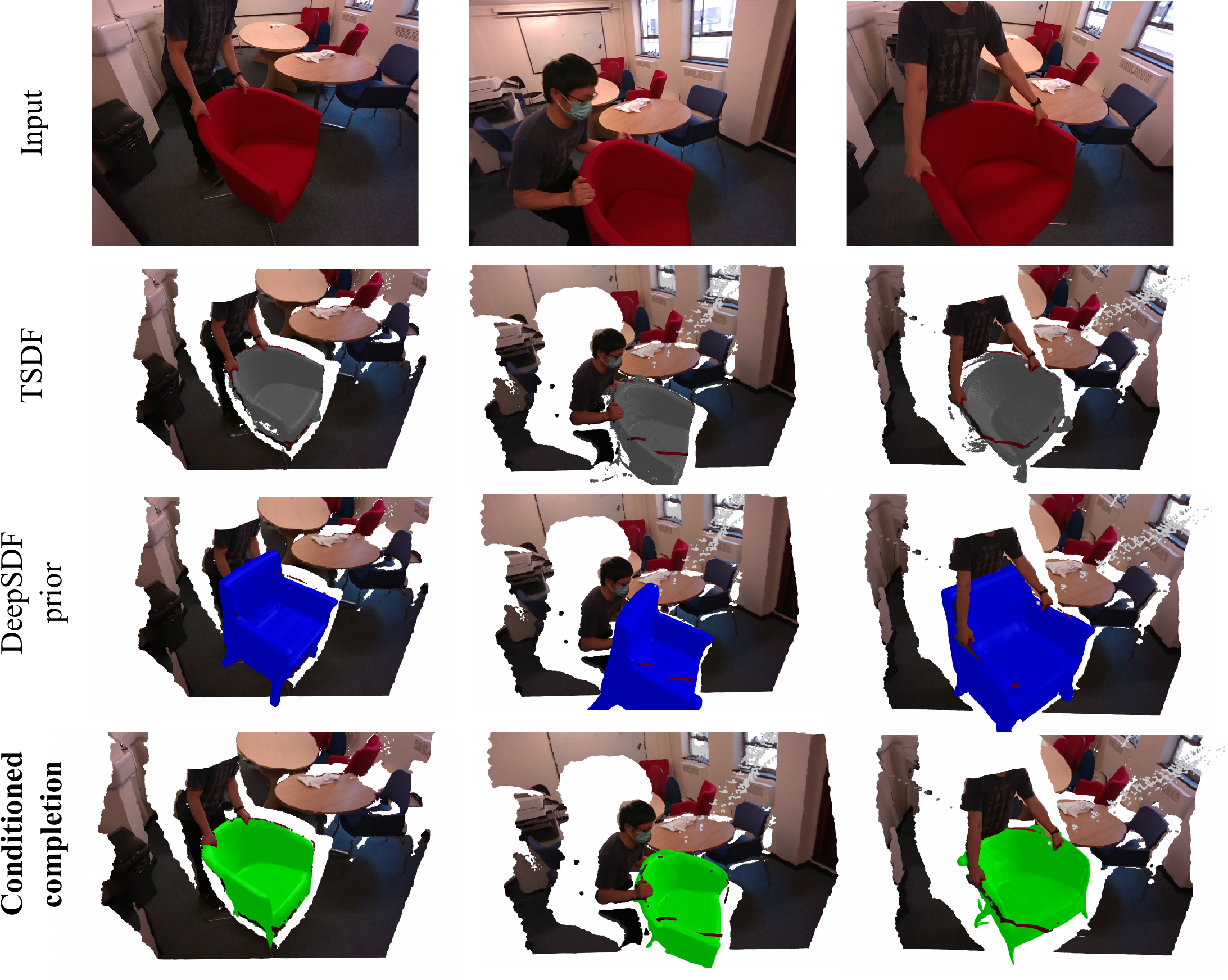}%
	\caption{Segment, track, reconstruct and complete a moving chair. Background pointclouds are just for visualization.}
	\label{fig:moving_chairs}
\end{figure}

We further demonstrate our proposed method in various real-world scenarios.  \cref{fig:shape_completion} shows the results in two different scenes. For each input image, we provide object reconstructions from the currently estimated camera viewpoint to visualise the observed part and from the top-down viewpoint to visualise the hidden part. As a qualitative comparison, we also show the reconstructions using classic TSDF fusion \cite{Newcombe:etal:ISMAR2011} and the learned category-level DeepSDF object prior \cite{Park:etal:CVPR2019}, it can be seen that TSDF-Fusion can only reconstruct the visible parts, leaving many empty holes in the object models. DeepSDF, on the other hand, has watertight reconstructions, but does not match the measurement necessarily, especially for the objects that deviate from the training space. On the contrary, our system can maintain highly detailed reconstructions and generate watertight meshes by filling in the missing parts using category-level shape priors thanks to the conditioned completion. \cref{fig:moving_chairs} also shows a scene where our system can reconstruct the visible parts and complete the hidden information of a moving object. The object pose and object geometry for the moving object are optimised jointly.

\section{Conclusions}
We present a novel approach for object-level tracking and mapping system in dynamic scenes by incorporating learned category-level shape priors. It enables to reasonably complete the object geometry of unseen parts based on the prior knowledge, and provide more robust and accurate tracking accuracy, even under large frame-by-frame motion and in dynamic environments with moving human involved. Experimental results in various scenarios demonstrate the effectiveness of our method. We hope our method can create a deeper understanding of exploring object prior information in object-level SLAM and benefit robots interacting with their surrounding environments. Continue from here, we would like to extend our system to a full graph-based object SLAM system.

\section*{ACKNOWLEDGMENT}
We thank Xingxing Zuo for fruitful discussions. This research is supported by Imperial College London, Technical University of Munich, and the EPSRC grant ORCA Stream B - Towards Resident Robots. Binbin Xu holds a China Scholarship Council-Imperial Scholarship.